\pdfoutput=1
\documentclass[journal]{IEEEtran}
\IEEEoverridecommandlockouts

\usepackage{booktabs}
\usepackage{cite}
\usepackage{subfigure}
\usepackage[ruled,vlined,algo2e]{algorithm2e}
\usepackage{algorithm}
\usepackage{algorithmic}
\usepackage{graphicx}
\usepackage{amssymb}
\usepackage{amsmath}
\usepackage{color}
\usepackage{graphicx}
\usepackage{subcaption}
\usepackage{url}
\usepackage[numbers, sort]{natbib}

\usepackage{makecell}
\usepackage{float}

\begin{document}
%

\title{A Wireless Foundation Model for Multi-Task Prediction}

\author{Yucheng Sheng,~\IEEEmembership{Student Member,~IEEE,}
        Jiacheng Wang,~\IEEEmembership{Student Member,~IEEE,}\\
        Xingyu Zhou,~\IEEEmembership{Student Member,~IEEE,}
        Le Liang,~\IEEEmembership{Member,~IEEE,}
        Hao Ye,~\IEEEmembership{Member,~IEEE,}\\
        Shi Jin,~\IEEEmembership{Fellow,~IEEE,}
        and Geoffrey Ye Li,~\IEEEmembership{Fellow,~IEEE}
        
\thanks{Yucheng Sheng, Jiacheng Wang, Xingyu Zhou, and Shi Jin are with the National Mobile Communications Research Laboratory, Southeast University, Nanjing 210096, China (e-mail: shengyucheng@seu.edu.cn; wangjiacheng@seu.edu.cn; \protect \url{xy_zhou@seu.edu.cn}; jinshi@seu.edu.cn).}
\thanks{Le Liang is with the National Mobile Communications Research Laboratory, Southeast University, Nanjing 210096, China, and also with Purple Mountain Laboratories, Nanjing 211111, China (e-mail: lliang@seu.edu.cn).}
\thanks{Hao Ye is with the Department of Electrical and Computer Engineering, University of California, Santa Cruz, CA 95064, USA (e-mail: hye30@ucsc.edu).}
\thanks{Geoffrey Ye Li is with the ITP Lab, the Department of Electrical and Electronic Engineering, Imperial College London, SW7 2BX London, U.K. (e-mail: geoffrey.li@imperial.ac.uk).}
}

\maketitle
\IEEEpeerreviewmaketitle
\begin{abstract}
With the growing complexity and dynamics of the mobile communication networks, accurately predicting key system parameters, such as channel state information (CSI), user location, and network traffic, has become essential for a wide range of physical (PHY)-layer and medium access control (MAC)-layer tasks. Although traditional deep learning (DL)-based methods have been widely applied to such prediction tasks, they often struggle to generalize across different scenarios and tasks. In response, we propose a unified foundation model for multi-task prediction in wireless networks that supports diverse prediction intervals. The proposed model enforces univariate decomposition to unify heterogeneous tasks, encodes granularity for interval awareness, and uses a causal Transformer backbone for accurate predictions. Additionally, we introduce a patch masking strategy during training to support arbitrary input lengths. After trained on large-scale datasets, the proposed foundation model demonstrates strong generalization to unseen scenarios and achieves zero-shot performance on new tasks that surpass traditional full-shot baselines.
\end{abstract}


\section{Introduction}
\IEEEPARstart{T}{he} advent of 6G communications \cite{hoydis20216g} has made wireless systems more intricate, featuring ultra-dense deployments, diverse service demands, and highly dynamic environments. Efficient execution of physical (PHY) and medium access control (MAC)-layer tasks require accurate and timely knowledge of the surrounding communication environment. Key parameters of interest include channel state information (CSI) \cite{jiang2019nnoverview}, user locations \cite{liu2022isaclstm}, mobile traffic at the base station (BS) \cite{chen2017lstmtraffic}, etc.  However, these parameters fluctuate rapidly over time, making real-time estimation and feedback particularly challenging. As a result, accurately predicting these variables has become essential for enabling a wide range of downstream communication tasks.

In this work, we focus on three representative prediction tasks spanning the PHY and MAC layers: channel prediction, angle prediction for integrated sensing and communication (ISAC), and traffic prediction for network management. 
Firstly, channel prediction in massive multiple-input multiple-output (MIMO) and orthogonal frequency division multiplexing (OFDM) systems has emerged as a promising approach to reduce the overhead of acquiring CSI. Both model-based and deep learning (DL)-based methods have been extensively studied. Model-based techniques, such as vector Prony-based prediction \cite{yin2020addressing} and high-resolution parameter estimation \cite{rottenberg2020performance}, provide accurate predictions under certain conditions. However, their reliance on theoretical models makes it challenging to ensure accuracy in practical channels across various scenarios. In contrast, DL-based methods, including recurrent neural networks (RNNs) \cite{jiang2019nnoverview}, long short-term memory (LSTM) networks \cite{jiang2020deep}, and Transformer models \cite{jiang2022transformer}, have been explored to capture the temporal dependencies in historical channel data for improved prediction accuracy. 
Secondly, most existing ISAC precoding design schemes typically rely on the assumption that the angular information of the sensing targets or communication users are known \cite{liu2021cramer}. In dynamic scenarios, such parameters are unavailable in real time. To address this issue, angle prediction can be employed to support precoding design and ensure reliable ISAC performance. Model-based approaches have been proposed to achieve high prediction accuracy, such as the extended Kalman filtering (EKF) method \cite{liu2020ekf} and Bayesian inference-based angular prediction algorithms \cite{yuan2020bayesian}. Additionally, DL-based models such as convolutional neural networks (CNNs) and LSTM networks in \cite{liu2022isaclstm} exploit spatial features and temporal dependencies from historical angle estimates, further enhancing prediction performance.
Thirdly, network management tasks, including load balancing \cite{liu2024spatial} and BS sleeping control \cite{lin2021data}, require proactive actions to anticipate future fluctuations in traffic demands \cite{bui2017survey}. As a result, traffic prediction plays a crucial role in the effective operation and management of communication networks. To address this challenge, both model-based methods, such as the autoregressive integrated moving average (ARIMA) model \cite{hendikawati2020survey}, and DL-based techniques, including LSTM networks \cite{huang2017ConvLSTM}, graph neural networks (GNNs) \cite{yu2017STGCN}, and attention-based models \cite{guo2019ASTGNN}, have been widely adopted for traffic forecasting.

The core challenge underlying the three aforementioned tasks, namely, channel, angle and traffic prediction, lies in accurately predicting future information based on historical data, situating these tasks within the broader field of time-series forecasting. DL-based approaches have outperformed model-based methods in the aforementioned tasks, demonstrating superior performance across various applications. However, despite their promising capabilities, DL-based time-series forecasting techniques still encounter several key challenges. A primary limitation is the restricted generalization ability of many DL-based models across diverse wireless environments, thereby limiting their effectiveness in real-world applications. In addition, most DL-based methods are exclusively designed for a single task, restricting their ability to handle multiple tasks simultaneously and limiting their adaptability and flexibility in dynamic, multi-faceted environments. Recently, large language models (LLMs) \cite{radford2019gpt}, \cite{touvron2023llama}  have demonstrated remarkable generalization and multi-task capabilities with their vast parameter sets, offering a potential solution to these issues.

Several studies have explored leveraging LLMs to enhance the performance of prediction tasks in wireless communications. For instance, prompt-as-prefix and patch embedding techniques were introduced in \cite{Sheng2025beam} that leverages LLM capabilities to improve the robustness of beam prediction. Similarly, to improve channel prediction accuracy and generalization, an LLM-driven approach for channel forecasting was proposed in \cite{liu2024llm4cp}. More recently, a multi-task PHY network driven by LLMs was developed in \cite{zheng2024large}, utilizing task-specific heads for different tasks. However, most of these LLM-based methods \cite{Sheng2025beam}, \cite{liu2024llm4cp} focus primarily on improving the generalization capability for individual tasks, lacking multi-tasking ability. The only exception in \cite{zheng2024large} intended for multi-task enhancement essentially relies on task-specific heads, which handle different tasks independently, rather than using a single unified model to address multiple tasks simultaneously. Moreover, all LLM-based approaches, whether utilizing GPT-2 \cite{radford2019gpt}, Llama-2 \cite{touvron2023llama}, or other variants, often store vast amounts of information irrelevant to communication tasks, leading to inefficiencies in model complexity and resource utilization. 
Recently, foundation models, which are pretrained on large-scale, domain-specific datasets and offer a unified architecture, have attracted significant attention due to their ability to excel in generalization and parameter efficiency \cite{bommasani2021opportunities}. These models show great potential in overcoming challenges across various prediction tasks in communication systems.


In response, we propose a unified foundation model for wireless prediction that supports diverse tasks across a wide range of time granularities in wireless networks. The main contributions of this paper are summarized as follows.
\begin{itemize}
    \item We propose a wireless foundation model for a range of prediction tasks in wireless networks. The model incorporates univariate decomposition techniques to unify diverse prediction tasks within a single framework and utilizes granularity encoding to classify input time granularities effectively. Additionally, the decoder-only framework leverages the causal attention mechanism, ensuring accurate and reliable predictions across all tasks. Specifically, this work focuses on three representative tasks, including channel prediction, angle prediction, and traffic prediction, while the proposed framework is flexible enough to accommodate other prediction tasks as well.
    \item For the pretraining of the proposed foundation model, we collected a large dataset encompassing various time intervals and multiple tasks. By employing the patch masking training method, we enable the foundation model to develop the ability to handle different history lengths effectively. Combined with univariate decomposition techniques, this approach ensures that the model can adapt to diverse data structures and generalize effectively across various prediction demands.
    \item Extensive experiments demonstrate that our proposed foundation model exhibits excellent prediction performance and remarkable generalization across scenarios and tasks. Regarding prediction performance, the predictions generated by the foundation model significantly outperform existing algorithms across all tasks. Experiments on downstream tasks further validate its performance. Regarding generalization ability, we first test it directly on a previously unseen heterogeneous channel prediction dataset, demonstrating the model's stable cross-scenario generalization. Then, we evaluate its performance on a new task, i.e., time-delay prediction in ISAC systems, and find that the model's zero-shot performance surpasses the full-shot performance of traditional algorithms. As the number of pretraining tasks increases, its zero-shot performance continues to improve. This highlights that the foundation model effectively captures the dynamics of wireless networks that are critical for time-series prediction, enabling strong performance across a wide range of wireless prediction tasks without requiring task-specific training.
\end{itemize}

The remainder of this paper is structured as follows. Section~\ref{sec:system_model} presents the system model and outlines the three selected tasks: channel prediction, angle prediction, and traffic prediction. Section~\ref{sec:foundation_model} provides a detailed explanation of the design and training process of the proposed foundation model. Simulation results are presented in Section~\ref{sec:performance evaluation}, and the paper concludes with Section~\ref{sec:conclusions}.


\section{System Model and Problem Formulation}\label{sec:system_model}

In this paper, we focus on multivariate time-series prediction in wireless networks, where the data is sampled at fixed intervals of $\Delta t$. Each time step $t$ is associated with an observation vector $\mathbf{x}_{t}\in \mathbb{R}^{M}$. Our goal is to predict all future values for the next $H$ time steps, using data from the previous $L$ time steps at time $t_0$. The time-series foundation model $f_{\theta}(\cdot)$, parameterized by $\theta$, predicts future values based on the historical context, given by
\begin{equation}
\mathbf{\hat{X}} = f_{\theta}(\mathbf{X};\Delta t), \label{eq:1}
\end{equation}
where $\mathbf{X}=[\mathbf{x}_{t_0-L+1}, \ldots, \mathbf{x}_{t_0} ]\in \mathbb{R}^{M \times L}$ represents the historical context of the previous $L$ time steps, and $\mathbf{\hat{X}}=[\hat{\mathbf{x}}_{t_0+1}, \ldots, \hat{\mathbf{x}}_{t_0+H}]\in \mathbb{R}^{M \times H}$ denotes the predicted values for the next $H$ time steps. 



\begin{figure}[htp]
    \centering
    \includegraphics[width=0.5\textwidth]{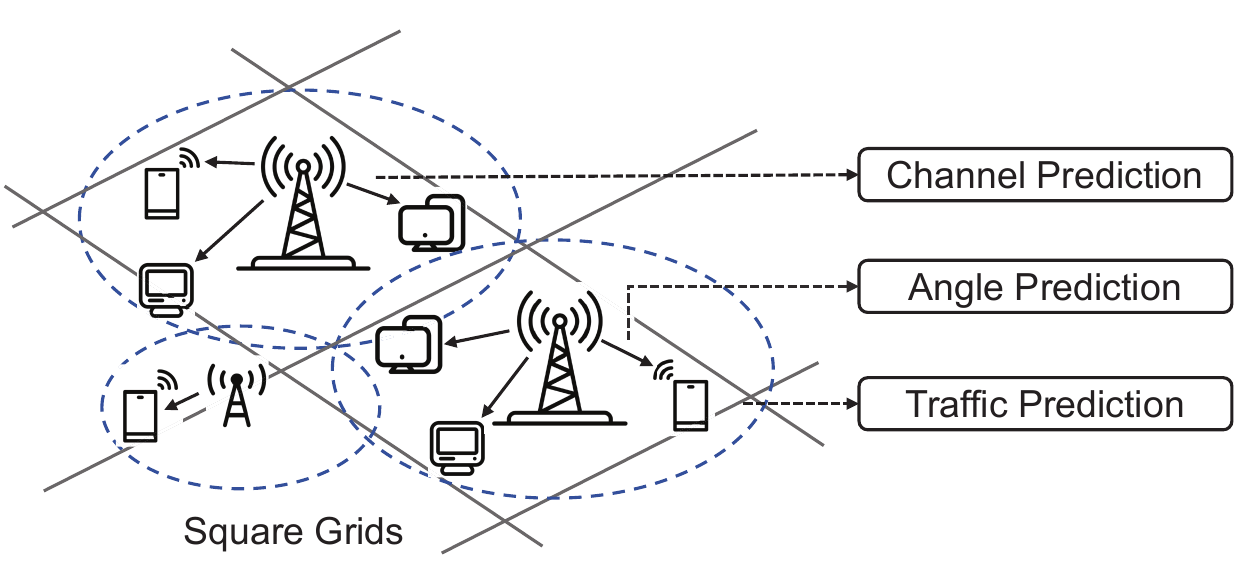}
    \caption{Three representative prediction tasks within wireless networks: channel prediction, angle prediction in ISAC systems, and traffic prediction.}
    \label{fig:prediction_task}
\end{figure}

We target three representative tasks across PHY and MAC layers: channel prediction, angle prediction in ISAC systems, and traffic prediction, as shown in Fig.~\ref{fig:prediction_task}. These tasks cover a wide range of time granularities, with prediction intervals $\Delta t$ spanning from milliseconds to hours. For each task, we utilize the proposed foundation model to predict key communication metrics, enabling more efficient and reliable network performance across a wide range of operating conditions.

\subsection{Channel Prediction}
We consider a multiple-input single-output (MISO) OFDM system working in a time-division duplexing (TDD) mode. The BS employs a uniform planar array (UPA) consisting of $N_t = N_h \times N_v$ antennas, where $N_h$ and $N_v$  are the number of antenna elements along the horizontal and vertical axes, respectively. The goal of this task is to predict the downlink CSI for the next $H$ time slots from historical CSI from the previous $L$ time slots, which can be acquired from uplink channel estimation with the help of the uplink-downlink CSI reciprocity in TDD systems. The CSI of $C$ subcarriers at $t$-th time slot can be represented as
\begin{equation}
\mathbf{H}_t=[\mathbf{h}_{1,t},\mathbf{h}_{2,t},\cdots,\mathbf{h}_{C,t}],\forall t,  \label{eq:csi}
\end{equation}
where $\mathbf{h}_{c,t} \in \mathbb{C}^{N_t}$ is the channel gain vector of subcarrier $c$ at $t$-th time slot.
To evaluate the channel prediction accuracy, the normalized mean-squared error (NMSE) between predicted CSI and ground-truth CSI is selected as the performance metric. Utilizing the metric, the channel prediction problem can be described as
\begin{align}
\min_{\theta} \quad
& \frac{
    \sum^{H}_{t=1} \left\| \hat{\mathbf{H}}_{t_0 + t} - \mathbf{H}_{t_0 + t} \right\|_F^2 \label{eq:channel_prediction}
}{
    \sum^{H}_{t=1} \left\| \mathbf{H}_{t_0 + t} \right\|_F^2
} \notag \\
\text{s.t.} \quad
& [ \hat{\mathbf{H}}_{t_0 + 1}, \ldots, \hat{\mathbf{H}}_{t_0 + H}]
= f_{\theta}\left( [\mathbf{H}_{t_0 - L + 1}, \ldots, \mathbf{H}_{t_0}] ; \Delta t \right),
\end{align}
where $\hat{\mathbf{H}}_{t}$ is the predicted CSI at time $t$ and $\left\| \cdot \right\|_F$ represents the Frobenius norm. We define the input and output sequences for channel prediction as $\mathbf{H}_{\text{CP}}=[\mathbf{H}_{t_0 - L + 1}, \ldots, \mathbf{H}_{t_0} ] \in  \mathbb{C}^{N_t \times C \times L }$ and $\mathbf{\hat{H}}_{\text{CP}}=[ \hat{\mathbf{H}}_{t_0 + 1}, \ldots, \hat{\mathbf{H}}_{t_0 + H}] \in  \mathbb{C}^{N_t \times C \times  H }$, respectively. Since the neural networks typically operate on real numbers, we convert the channel matrices $\mathbf{H}_{\text{CP}}$ and $\mathbf{\hat{H}}_{\text{CP}}$ into real tensors: $\mathbf{X}_{\text{CP}} \in \mathbb{R}^{2N_t C \times L}$ and  $\mathbf{\hat{X}}_{\text{CP}} \in \mathbb{R}^{2N_t C \times H}$, where the channel dimension $2N_t C$ is subsumed into the multivariate dimension $M$. 

Unlike prior work \cite{liu2024llm4cp}, \cite{zheng2024large}, where the prediction interval is typically within the small-scale fading range, we consider two distinct scenarios: $\Delta t = 0.5$ ms and $\Delta t = 50$ ms. The latter attempts to approach the time scale of large-scale fading, which is often influenced by path loss and shadowing, leading to the larger divergence across time. Therefore, for $\Delta t = 50$ ms, we apply transformations such as $\log_{10}(\mathbf{X}_{\text{CP}})$ and $\log(\hat{\mathbf{X}}_{\text{CP}})$ to handle the larger dynamic range.

\textbf{Downstream Task:}
To evaluate the practical utility of the predicted CSI, we consider a downlink MISO-OFDM transmission scenario utilizing $C$ subcarriers. To maximize spectrum efficiency, we adopt matched-filtering based precoding as
\begin{equation}
\mathbf{w}_{c,t} = \frac{\mathbf{\hat{h}}_{c,t}}{\|\mathbf{\hat{h}}_{c,t}\|},
\end{equation}
where $\mathbf{\hat{h}}_{c,t}$ denotes predicted CSI. The received downlink signal of the $c$-th subcarrier is given by
\begin{equation}
y_{c,t} = \mathbf{h}_{c,t}^H \mathbf{w}_{c,t} x_{c,t} + n_{c,t},
\end{equation}
where $x_{c,t}$ is the transmitted signal, and $n_{c,t}$ is the additive white Gaussian noise (AWGN) with noise power $\sigma_n^2$. Therefore, the spectrum efficiency of the downlink transmission process is derived as
\begin{equation}
R_{\text{CP}} = \frac{1}{HC}\sum_{c=1}^{C}  \sum_{t=t_0 + 1}^{t_0 + H} \log_2 \left( 1 + \frac{|\mathbf{h}_{c,t}^H \mathbf{w}_{c,t}|^2}{\sigma_n^2} \right). \label{eq:se}
\end{equation}
It is important to note that prediction inaccuracies in $\mathbf{\hat{h}}_{c,t}$ can lead to suboptimal $\mathbf{w}_{c,t}$, thus degrading the overall spectrum efficiency. This evaluation serves to quantify how prediction quality translates into practical communication performance.

\subsection{Angle Prediction}
In ISAC systems, it is important to allocate resources for communication and sensing in the spatial domain, specifically through beamforming, with allocation based on the user's angle information. In real-world mobile scenarios, predicted angle information can be used as prior knowledge to assist in the design of beam vectors.
In this task, we examine an ISAC-enabled system, where a monostatic BS simultaneously communicates with $K$ single-antenna users while sensing their locations. The BS, equipped with a millimeter-wave (mmWave) uniform linear array (ULA), operates at full-duplex mode, comprising $N_t$ transmit antennas and $N_r$ receive antennas. At each time instant $\tau$ within the $t$-th time interval, the ISAC transmit signal vector is represented as
\begin{equation}
\mathbf{x}_t(\tau) = \mathbf{W}_t \mathbf{s}_t(\tau) \in \mathbb{C}^{N_t },
\end{equation}
where $\mathbf{W}_t = [\mathbf{w}_{1,t}, \ldots, \mathbf{w}_{K,t}] \in \mathbb{C}^{N_t \times K}$ represents the transmit beamforming matrix, and $\mathbf{w}_{k,t} \in \mathbb{C}^{N_t }$ denotes the beamforming vector with respect to the $k$-th user. Accordingly, the BS receives the echo signals reflected from the users~\cite{liu2020ekf}, which is given by
\begin{equation}
\mathbf{r}_t(\tau) = G \sum_{k=1}^{K} \beta_{k,t} e^{j2\pi \mu_{k,t} \tau} \mathbf{b}(\phi_{k,t}) \mathbf{a}^H(\phi_{k,t}) \mathbf{x}_t(\tau - \nu_{k,t}) + \mathbf{z}(\tau), \label{eq:isac}
\end{equation}
where $\nu_{k,t}$ and $\mu_{k,t}$ denote the time-delay and Doppler shift of the $k$-th user, respectively. The overall antenna array gain is given by $G = \sqrt{N_t N_r}$. The angle of arrival (AoA) for the $k$-th user at time slot $n$ is denoted as $\phi_{k,n}$, while the reflection coefficient is defined as $\beta_{k,n} = {\rho}/{2d_{k,n}}$, where $\rho$ represents the fading factor and $d_{k,n}$ is the distance between the user and the BS at time slot $t$. Moreover, $\mathbf{z}(\tau) \in \mathbb{C}^{N_r}$ represents the noise received in the BS. A line-of-sight (LOS) channel model~\cite{niu2015survey} is generally adopted for mmWave systems, where the transmit and receive beamforming vectors in the BS are represented as
\begin{equation}
\mathbf{a}(\phi_{k,t}) = \sqrt{\frac{1}{N_t}} \left[1, e^{-j\pi \cos \phi_{k,t}}, \cdots, e^{-j\pi (N_t - 1) \cos \phi_{k,t}} \right]^{\text{T}},
\end{equation}
and
\begin{equation}
\mathbf{b}(\phi_{k,t}) = \sqrt{\frac{1}{N_r}} \left[1, e^{-j\pi \cos \phi_{k,t}}, \cdots, e^{-j\pi (N_r - 1) \cos \phi_{k,t}} \right]^{\text{T}},
\end{equation}
respectively. Utilizing existing angle estimation methods to the received echo signals in (\ref{eq:isac}), we can acquire the historical angle vector at time slot $t$, expressed as $\boldsymbol{\phi}_{t} = \left[ \phi_{1,t}, \ldots, \phi_{K,t} \right]^{\text{T}} \in \mathbb{R}^{K}$. In a manner analogous to \eqref{eq:channel_prediction}, the angle prediction task can be formulated as
\begin{align}
\min_{\theta} \quad
& \frac{
    \sum^{H}_{t=1} \left\| \boldsymbol{\hat{\phi}}_{t_0 + t} - \boldsymbol{\phi}_{t_0 + t} \right\|_2^2 \label{eq:angle_prediction}
}{
    \sum^{H}_{t=1} \left\| \boldsymbol{\phi}_{t_0 + t} \right\|_2^2
} \notag \\
\text{s.t.} \quad
& [ \boldsymbol{\hat{\phi}}_{t_0 + 1}, \ldots, \boldsymbol{\hat{\phi}}_{t_0 + H}]
= f_{\theta}\left( [ \boldsymbol{\phi}_{t_0 - L + 1}, \ldots, \boldsymbol{\phi}_{t_0}] ; \Delta t \right), 
\end{align}
where $\boldsymbol{\hat{\phi}}_{t}$ is the predicted angles at the $t$-th time slot. Similarly, we directly unify the input and output sequence for angle prediction into the expression of $\mathbf{X}_{\text{AP}}=[\boldsymbol{\phi}_{t_0-L+1},\ldots,\boldsymbol{\phi}_{t_0}] \in \mathbb{R}^{K \times L}$ and $\hat{\mathbf{X}}_{\text{AP}}=[\boldsymbol{\hat{\phi}}_{t_0+1},\ldots,\boldsymbol{\hat{\phi}}_{t_0+H}]  \in \mathbb{R}^{K \times H}$, where the number of users $K$ is subsumed into the multivariate dimension $M$.

\textbf{Downstream Task:}
In this task, we assess the utility of angle prediction within an ISAC system by examining its impact on downlink beamforming and user signal quality. The goal is to evaluate how the predicted AoA at the BS influences the received signal quality and the achievable data rate for each user. Specifically, the received signal at the $k$-th user can be expressed as
\begin{equation}
\vartheta_{k,t}(\tau) = \tilde{G} \sqrt{\alpha_{k,t}} e^{j 2 \pi f_k t \tau} \mathbf{a}^H(\phi_{k,t}) \sum_{i=1}^{K} \mathbf{w}_{i,t} s_{i,t}(\tau) + n_{k,t}(\tau),
\end{equation}
where $\vartheta_{k,t}(\tau)$ denotes the received signal at the $\tau$-th time instant within the $t$-th time slot, and $\tilde{G} = \sqrt{N_t}$ represents the antenna gain. Additionally, the path loss is modeled as $\alpha_{k,t} = \alpha_0 \left( \frac{d_{k,t}}{d_0} \right)^{-\zeta}$, where $\alpha_0$ is the reference path loss at distance $d_0$, and $\zeta$ is the path loss exponent. The additive noise at the $k$-th user is denoted by $n_{k,t}(\tau) \sim \mathcal{CN}(0, \sigma_k^2)$, with $\sigma_k^2$ indicating the noise variance. Under this model, the signal-to-interference-plus-noise ratio (SINR) received at the $k$-th user is represented by
\begin{equation}
\text{SINR}_{k,t} = \frac{|\tilde{G} \sqrt{\alpha_{k,t}} \mathbf{a}^H(\phi_{k,t}) \mathbf{w}_{k,t}|^2}
{\sum_{j \in \mathcal{K}, j \ne k} |\tilde{G} \sqrt{\alpha_{k,t}} \mathbf{a}^H(\phi_{k,t}) \mathbf{w}_{j,t}|^2 + \sigma_k^2}.
\end{equation}

Since massive MIMO systems employ a large number of transmit antennas, the beamforming vectors corresponding to distinct angles can be considered asymptotically orthogonal. That is, for any $k \ne k'$, we have $\left|\mathbf{a}^H(\phi_{k,t}) \mathbf{a}(\phi_{k',t})\right| \approx 0.$ Under this assumption, if the beamforming vector is designed as $\mathbf{w}_{k,t} = \sqrt{p_{k,t}} \mathbf{a}(\hat{\phi}_{k,t})$, where $p_{k,t}$ denotes the transmit power and $\hat{\phi}_{k,t}$ is the estimated angle between the $k$-th user and the BS, the inter-user interference becomes negligible. As a result, the SINR reduces to the signal-to-noise ratio (SNR), which can be expressed as
\begin{equation}
\text{SNR}_{k,t}(\hat{\phi}_{k,t}) = \frac{p_{k,t} |\tilde{G} \sqrt{\alpha_{k,t}} \mathbf{a}^H(\phi_{k,t}) \mathbf{a}(\hat{\phi}_{k,t})|^2}
{\sigma_k^2}.
\end{equation}

Correspondingly, the downlink sum spectrum efficiency, averaged over $H$ time slots, can be formulated as
\begin{equation}
R_{\text{AP}} = \frac{1}{H}\sum_{k=1}^{K} \sum_{t=t_0 + 1}^{t_0 + H} \log_2 \left(1 + \text{SNR}_{k,t}(\hat{\phi}_{k,t})\right).  \label{eq:sum_rate_isac}
\end{equation}

\subsection{Traffic Prediction}
In this task, we focus on Internet traffic prediction within an area consisting of $N_1\times N_2$ square grids, with traffic in each grid cell aggregated on an hourly basis. Similarly, the traffic prediction problem is formulated as
\begin{align}
\min_{\theta} \quad
& \frac{
\sum^{H}_{t=1} \left\| \hat{\mathbf{I}}_{t_0 + t} - \mathbf{I}_{t_0 + t} \right\|_F^2 \label{eq:traffic_prediction}
}{
\sum^{H}_{t=1} \left\| \mathbf{I}_{t_0 + t} \right\|_F^2
} \notag \\
\text{s.t.} \quad
& [ \hat{\mathbf{I}}_{t_0 + 1}, \ldots, \hat{\mathbf{I}}_{t_0 + H}]
= f_{\theta}\left( [\mathbf{I}_{t_0 - L + 1}, \ldots, \mathbf{I}_{t_0}] ; \Delta t \right),
\end{align}
where $\mathbf{I}_{t} \in \mathbb{R}^{N_1\times N_2}$ denotes the actual Internet traffic at the $t$-th time step across all grids, while $\hat{\mathbf{I}}_{t}$ represents the predicted Internet traffic. Finally, we transform the input and output matrix sequence for traffic prediction into tensors of $\mathbf{X}_{\text{TP}} \in \mathbb{R}^{N_1N_2 \times L}$ and $\hat{\mathbf{X}}_{\text{TP}} \in \mathbb{R}^{N_1N_2 \times H}$, where the number of squares $N_1N_2$ is subsumed into the multivariate dimension $M$.


\section{Time-Series Foundation Model}\label{sec:foundation_model}

In this section, we present the design of the proposed wireless foundation model for multi-task prediction. We begin by describing the data processing pipeline, which converts raw inputs from various tasks into a unified format suitable for pretraining. Next, we elaborate on the model architecture, emphasizing key components including input embedding, positional encoding, granularity encoding, and the causal Transformer backbone. Finally, we introduce the learning algorithm, detailing the patch masking strategy and the loss function employed during training.

\begin{figure*}[tbp]
\centering
\includegraphics[scale=0.65]{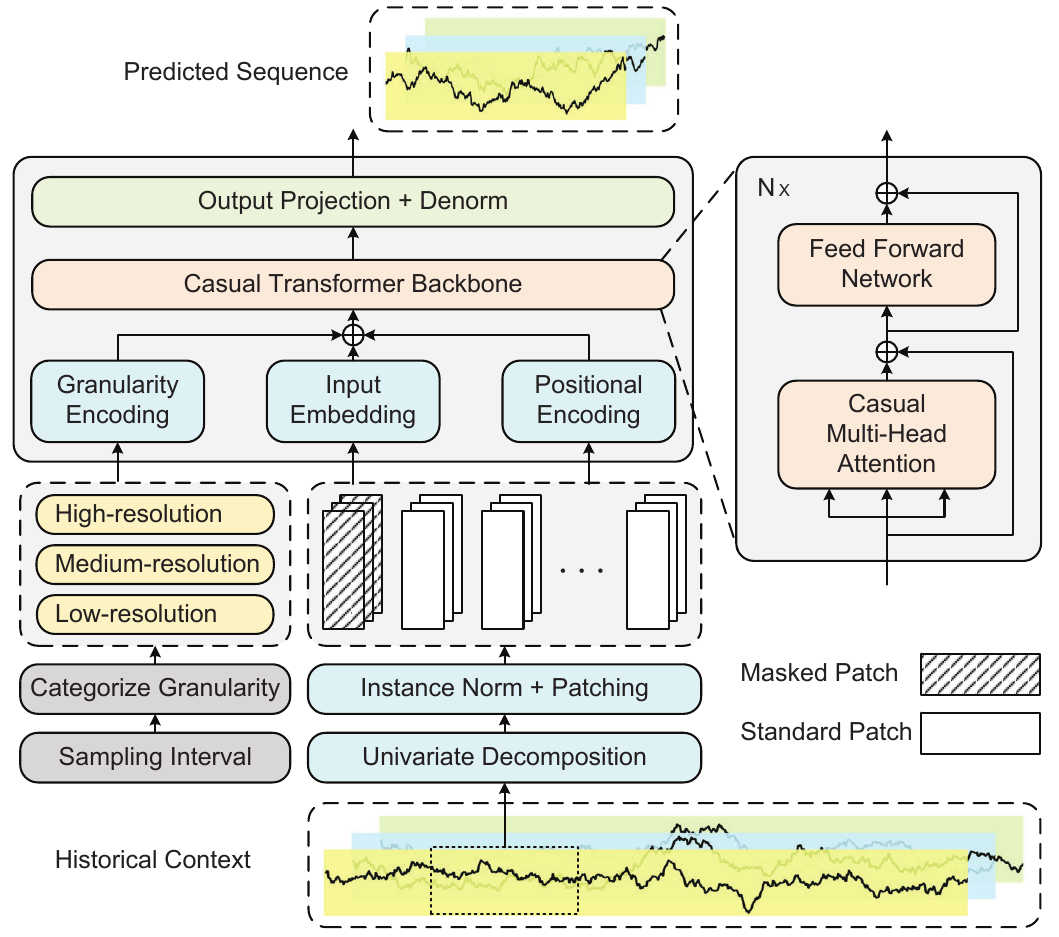}
\caption{The architecture of the proposed wireless foundation model for multi-task prediction.}
\label{fig:TSFM_model}
\end{figure*}
\subsection{Data Processing}
\textbf{Univariate Decomposition.}
The paper \cite{nie2022time} suggests that processing each channel separately in multivariate time-series data often outperforms the technique of mixing multi-channel data. Inspired by this, we apply the univariate decomposition technique to split $\mathbf{X} \in \mathbb{R}^{M \times L }$ into $M$ univariate time-series, represented as
\begin{equation}
{\left[\mathbf{x}_{\text{his}}^{(1)},\ldots,\mathbf{x}_{\text{his}}^{(M)} \right]}^\top = \text{RowSplit}\left(\mathbf{X}\right),
\end{equation}
where $\mathbf{x}_{\text{his}}^{(m)} = [x_{t_0-L+1}^{(m)}, \ldots, x_{t_0}^{(m)}]\in \mathbb{R}^{L}, \forall m \in \{1, \dots, M\}$ represents the historical sequence in the $m$-th channel of the original multivariate time-series. Each of these univariate series is processed independently using a shared Transformer architecture, but with separate forward passes.



\textbf{Instance Normalization.}
To mitigate the distribution shift in time-series and improve the model robustness, we apply instance normalization \cite{ulyanov2016instance} to each univariate time-series $\mathbf{x}_{\text{his}}^{(m)}$ during both the training and testing phases, given by
\begin{equation}
\mathbf{x}_{\text{norm}}^{(m)} = \text{InstanceNorm}\left(\mathbf{x}_{\text{his}}^{(m)}\right),
\end{equation}
where the normalizated series $\mathbf{x}_{\text{norm}}^{(m)} \in \mathbb{R}^{L}$ has zero mean and unit variance. The mean and standard deviation will be reintroduced to the output prediction $\hat{\mathbf{x}}^{(m)}$.

\textbf{Patching.} 
Time-series sequences often have significantly longer input lengths than those typically seen in natural language processing (NLP) tasks. Therefore, we use the patching method defined in \cite{nie2022time} to process $\mathbf{x}_{\text{norm}}^{(m)}$ to improve model adaptability, given by
\begin{equation}
\mathbf{X}_{\text{pat}}^{(m)} = \text{Patching}\left(\mathbf{x}_{\text{norm}}^{(m)}\right),
\end{equation}
where $\mathbf{X}_{\text{pat}}^{(m)} = [\mathbf{x}_{1,\text{pat}}^{(m)},\ldots,\mathbf{x}_{N_p,\text{pat}}^{(m)}]^\top \in \mathbb{R}^{N_p \times L_p}$ represents the contiguous, non-overlapping patches after patching, $L_p$ is the length of each patch, and $N_p=\lceil L/L_p \rceil$ is number of patches. These patches are similar to tokens in language models, capturing local semantic information within the time-series. Additionally, by reducing the number of tokens input to the Transformer by a factor of $L_p$, this technique significantly reduces the time and space complexity.

\subsection{Model Architecture}

As shown in Fig.~\ref{fig:TSFM_model}, the multivariate historical context is first decomposed into univariate patches through univariate decomposition, instance normalization, and patching, making them suitable for processing by the foundation model. These patches are then mapped to a high-dimensional latent space through input embedding and positional encoding. Additionally, the sampling interval is extracted as time granularity, and its corresponding granularity encoding is combined with the aforementioned embedding before being input into the decoder-only causal Transformer backbone. After feature extraction by the stacked Transformer, the final features are used to directly predict the required sequence through output projection. Furthermore, patch masking will be applied to the first patch of each historical sequence during the pretraining process.

\textbf{Input Embedding.}
To extract semantic features from each input patch, we apply a residual network (ResNet) \cite{he2016deep} to project the patch sequence into a high-dimensional latent space. For the $m$-th univariate input sequence, the patch embedding is given by:
\begin{equation}\label{eq:input_embedding}
\mathbf{E}_{\text{pat}}^{(m)} = \text{ResNet}\left(\mathbf{X}_{\text{pat}}^{(m)}\right),
\end{equation}
where $\mathbf{E}_{\text{pat}}^{(m)} \in \mathbb{R}^{N_p \times d_m}$ represents the input embedding, and $d_m$ is the dimension of the Transformer latent space.

\textbf{Positional Encoding.}
To preserve the temporal order of patches that are essential for sequence modeling, we add positional encodings to the patch embeddings before feeding them into the Transformer. This positional encoding is computed as
\begin{equation}
\mathbf{E}_{\text{pos}}^{(m)} = \text{PE}\left(\mathbf{j}\right),
\end{equation}
where $\mathbf{E}_{\text{pos}}^{(m)}\in \mathbb{R}^{N_p \times d_m}$ denotes the position encoding, PE denotes the fixed position encoding function in \cite{vaswani2017attention}, and $\mathbf{j} \in \mathbb{R}^{L_{p}}$ represents the index vector of the patch positions.

\textbf{Granularity Encoding.}
We notice that the inherent pattern of the time-series changes with the scale of the sampling interval, and the pattern difference increases as the scale difference grows, as discussed in the ablation studies in Section \ref{sec:ablation studies}. Inspired by this, we categorize the time granularity $g$ into three resolution levels based on the sampling interval $\Delta t$ as
\begin{equation}
    g = \begin{cases}
\text{high}, & \text{if } \Delta t < 1 \, \text{ms}, \\
\text{medium}, & \text{if } 1 \, \text{ms} \leq \Delta t < 1 \, \text{h}, \\
\text{low}, & \text{if } \Delta t \geq 1 \, \text{h}.
\end{cases}
\end{equation}
We introduce granularity encoding to provide resolution information to the model, expressed as
\begin{equation}
\mathbf{E}_{\text{gra}}^{(m)} = \text{GE}\left(g\right),
\end{equation}
where $\mathbf{E}_{\text{gra}}^{(m)}\in \mathbb{R}^{N_p \times d_m}$ denotes the granularity encoding, and GE represents a trainable lookup table for each time granularity $g$. Finally, combining the input embedding, positional encoding and granularity encoding, we obtain the Transformer input as follows
\begin{equation}
\mathbf{E}^{(m)} = \mathbf{E}_{\text{pat}}^{(m)} + \mathbf{E}_{\text{pos}}^{(m)} +  \mathbf{E}_{\text{gra}}^{(m)}.
\end{equation}

\textbf{Causal Transformer Backbone.}
We use a causal Transformer to extract features from the input embeddings, consisting of a stack of Transformer layers. Each Transformer layer contains a multi-head causal self-attention mechanism followed by a feed-forward network. For each head $h = \{1,\cdots, H\}$ in the multi-head attention mechanism, we transform $\mathbf{E}^{(m)}$ into a query matrix $\mathbf{Q}^{(m)}_h=\mathbf{E}^{(m)}\mathbf{W}^Q_h$, a key matrix $\mathbf{K}^{(m)}_h=\mathbf{E}^{(m)}\mathbf{W}^K_h$, and a value matrix $\mathbf{V}^{(m)}_h=\mathbf{E}^{(m)}\mathbf{W}^V_h$, where $\mathbf{W}^Q_h,\mathbf{W}^K_h\in \mathbb{R}^{d_m \times d_k}$ and $\mathbf{W}^V_h\in \mathbb{R}^{d_m \times d_v}$. After that, the $h$-th head after causal attention is given by
\begin{equation}
\text{Head}^{(m)}_h = \text{Softmax}\left(\frac{\mathbf{Q}^{(m)}_h \left(\mathbf{K}^{(m)}_h \right)^{\top} + \mathbf{M}}{\sqrt{d_k}}\right)\mathbf{V}^{(m)}_h,
\end{equation}
where $\mathbf{M}\in \mathbb{R}^{N_p \times N_p}$ represents the masking matrix for causal attention with each element, given by
\begin{equation}
\mathbf{M}[i,j] = 
\begin{cases} 
0, & \text{if } i \geq j,\\
-\infty, & \text{if } i < j,
\end{cases}
\end{equation}
which ensures that each position $i$ can only attend to itself and previous positions $j \leq i$, while future positions $j > i$ are masked out with $-\infty$. Using the multi-head attention mechanism, the model attends to information from different representation subspaces simultaneously, thereby obtaining
\begin{equation}
\mathbf{O}^{(m)} = \text{Concat}\left(\text{Head}^{(m)}_1,\ldots,\text{Head}^{(m)}_H\right)\mathbf{W}^O,
\end{equation}
where $\mathbf{O}^{(m)}\in \mathbb{R}^{N_p \times d_m}$ represents the output of multi-head causal attention, and $\mathbf{W}^O\in \mathbb{R}^{H d_v \times d_m}$ linearly projects these heads to align with the latent dimension of the Transformer. The attention output $\mathbf{O}^{(m)}$ passes through the subsequent feed-forward network and other stacked Transformer layers in the backbone, and is finally converted into a feature matrix $\mathbf{Z}^{(m)} \in \mathbb{R}^{N_p \times d_m}$.

\textbf{Output Projection.}
We employ a decoder-only architecture for time-series prediction, where all input patches are utilized to predict future values beyond the final input time step. Unlike language models, our time-series foundation model allows the output sequence length $H$ to differ from the patch length $L_p$. To produce the output, we incorporate a ResNet module that directly generates the complete $H$-step prediction. Importantly, the final patch representation $\mathbf{z}^{(m)}_{N_p} \in \mathbb{R}^{d_m}$ encodes global information aggregated from all preceding patches,  due to the design of the feature extraction mechanism. Therefore, to ensure robustness to the varying number of patches $N_p$, we rely solely on $\mathbf{z}^{(m)}_{N_p}$ for output generation, computed as
\begin{equation}
\hat{\mathbf{x}}_{\text{norm}}^{(m)} = \text{ResNet}\left(\mathbf{z}^{(m)}_{N_p}\right),
\end{equation}
where $\hat{\mathbf{x}}_{\text{norm}}^{(m)} \in \mathbb{R}^{H}$ represents the normalized prediction result, and the prediction result $\hat{\mathbf{x}}_\text{fut}^{(m)}\in \mathbb{R}^{H}$ is obtained after inverse instance normalization. This approach avoids multi-step autoregressive iterations, reducing the number of iterations required for model inference and significantly improving performance \cite{zeng2023transformers}.

\subsection{Learning Algorithm}

\textbf{Patch Masking.}
While patching enables efficient tokenization of time-series data, it introduces a constraint: the history length $L$ must be divisible by the patch length $L_p$ during inference. To ensure the model generalizes to arbitrary history lengths $L$, we introduce patch masking \cite{das2024decoder} during training. Let $\mathbf{x}_{j,\text{pat}}^{(m)} \in \mathbb{R}^{L_p}$ denote the $j$-th patch of the $m$-th univariate sequence, we then mask this patch as:
\begin{equation}
\mathbf{x}_{j,\text{mask}}^{(m)} = \mathbf{x}_{j,\text{pat}}^{(m)} \odot \mathbf{m}_{j,\text{pat}}^{(m)},
\end{equation}
where $\odot$ denotes the Hadamard product, and $\mathbf{m}_{j,\text{pat}}^{(m)} \in \mathbb{R}^{L_p}$ is the $j$-th patch mask. In the training phase, given history length $L=N_p \times L_p$, we mask only the first patch, i.e., $\mathbf{m}_{j,p}^{(m)} = \mathbf{1},\forall j>1$. For the first patch, we set $\mathbf{m}_{j,p}^{(m)} = [0,\ldots,1]^\top$, with the first $r$ elements set to 0 and the remaining parts to 1, where $r$ is a random integer selected between 0 and $L_p-1$. After patch masking, we concatenate these masked patches into $\mathbf{X}_{\text{mask}}^{(m)} = [\mathbf{x}_{1,\text{mask}}^{(m)}, \ldots, \mathbf{x}_{N_p,\text{mask}}^{(m)}]^\top \in \mathbb{R}^{N_p \times L_p}$, and replace $\mathbf{X}_{\text{pat}}^{(m)}$ in \eqref{eq:input_embedding}. Iterating over all possible input patch numbers $N_p$, the model supports any input history lengths $L$. In the inference phase, we prepend zeros to the input sequence until $L$ becomes an integer multiple of $L_p$ before patching.

\textbf{Loss Function.}
We use the mean-squared error (MSE) loss to measure the difference between the prediction $\hat{\mathbf{x}}_\text{fut}^{(m)}$ and the true value $\mathbf{x}_{\text{fut}}^{(m)} = [x_{t_0+1}^{(m)}, \ldots, x_{t_0+H}^{(m)}]\in \mathbb{R}^{H}, \forall m \in \{1, \dots, M\}$. The overall loss is defined as
\begin{equation}
\mathcal{L}=\frac{1}{MH} \sum_{m=1}^{M} \text{MSE}\left(\hat{\mathbf{x}}_\text{fut}^{(m)}, \mathbf{x}_\text{fut}^{(m)}\right),
\end{equation}
which averages the loss over all $M$ variables and $H$ predictions, ensuring balanced optimization across all dimensions.



\section{Performance Evaluation} \label{sec:performance evaluation}

In this section, we first present the experimental setup and datasets used to train and evaluate the proposed foundation model. For each task, we provide a detailed description of the baseline methods, performance metrics, and a comprehensive analysis of the results. We further assess the effectiveness of the proposed foundation model on corresponding downstream tasks. In addition, we examine the model’s zero-shot generalization capabilities, evaluate its robustness under varying conditions, and conduct ablation studies to quantify the contribution of each component within the framework.

\subsection{Simulation Setup and Datasets}
\textbf{Simulation Setup}:
During training, the model was configured with 16 stacked causal Transformer layers. The maximum history lengths for channel prediction, angle prediction, and traffic prediction were set to 32, 32, and 64, respectively, while the prediction length for all three tasks was fixed at 4. Unless otherwise noted, these settings are adopted throughout all subsequent experiments. The training was conducted with a batch size of 2,048 and a learning rate of $1 \times 10^{-5}$. All experiments were performed on a single machine equipped with two NVIDIA GeForce RTX 4090 GPUs over a period of five days.

\textbf{Dataset for Channel Prediction}: In order to fully exploit the prediction capabilities of foundation models across different CSI configurations, we have created a set of diverse CSI datasets, generated using the QuaDRiGa channel model \cite{quadriga2014}, adhering to 3GPP standards. We focus on MISO-OFDM systems, where the BS uses a UPA with $N_h = 4$ elements horizontally and $N_v = 4$ elements vertically, while the users are equipped with a single antenna. The antenna spacing is set to half of the wavelength at the center frequency. We assume a TDD system with a center frequency of 2.4 GHz. The channel has a bandwidth of 8.64 MHz and consists of $C = 48$ subcarriers, resulting in a 180 kHz subcarrier spacing. Each CSI sample consists of a sequence of CSI observations over time, generated as the user moves along a straight-line path starting from a random position, with the speed chosen randomly within the range of 10 to 100 kilometer per hour. As summarized in Table \ref{tab:csi_configuration}, we simulate 11 datasets, labeled D1 to D11, covering various frequencies, channel models, propagation environments (LOS or non-line-of-sight (NLOS)), and user speeds. Datasets D1 through D10 are used for pretraining, while D11 is reserved for evaluating the model's capability of generalization to unseen conditions. The pretraining dataset is partitioned into training and testing subsets, containing 12,960,000 and 3,240,000 samples, respectively. 
\begin{table}[t]
\centering
\caption{An illustration of the system configurations of the constructed CSI datasets.}
\begin{tabular}{c|ccc}
\Xhline{0.8pt}
Dataset & $\Delta t$ (ms) & Channel Model & Scenarios \\
\Xhline{0.4pt}
D1 & 0.5 & UMi & LOS \\
D2 &  0.5 & UMi &  NLOS \\
D3 &  0.5 & UMa &   LOS\\
D4 & 0.5& UMa & NLOS \\
D5 & 0.5  & RMa &   NLOS\\
D6 & 50 & UMi &  LOS\\
D7 &  50& UMi &  NLOS\\
D8 & 50 & UMa & LOS \\
D9 & 50 & UMa &   NLOS\\
D10 & 50 & RMa & NLOS \\
\Xhline{0.4pt}
D11 & 0.5 & RMa & LOS \\
\Xhline{0.8pt}
\end{tabular}
\label{tab:csi_configuration}
\end{table}

\textbf{Dataset for Angle Prediction}: In this task, we define four mobility modes for the dataset: pedestrian, bicycle, vehicle, and rural vehicle, with corresponding speed ranges of 0-10 km/h, 10–25 km/h, 30–80 km/h, and 80–120 km/h, respectively. The sampling intervals are set to $\Delta t = 50~\text{or}~100$ ms. The distance between the BS and the user varies between 5 and 500 meters. Additionally, the user experiences random acceleration and changing directions during motion. The pretraining dataset is partitioned into training and testing subsets, containing 6,336,000 and 1,584,000 samples, respectively. 

\textbf{Dataset for Traffic Prediction}: 
We use a publicly available multi-source dataset released by Telecom Italia in 2015 \cite{barlacchi2015multi}, which is recognized as one of the largest operator-provided datasets to date. Specifically, we concentrate on the mobile Internet usage data from the city of Milan, which is spatially partitioned into 100 $\times$ 100 square grids, each covering an area of approximately 0.235 $\times$ 0.235 $\text{km}^2$. Each grid cell contains aggregated Internet traffic. The entire grid coverage is utilized for pretraining purposes.  The sampling interval is set to $\Delta t = 1$ h.  The pretraining data is subsequently divided into training and testing subsets, comprising 6,389,000 and 172,000 samples, respectively.

\begin{figure*}[htp]
	\centering
	\subfigure[$\Delta t = 0.5$  ms]{
		\includegraphics[width=0.48\textwidth]{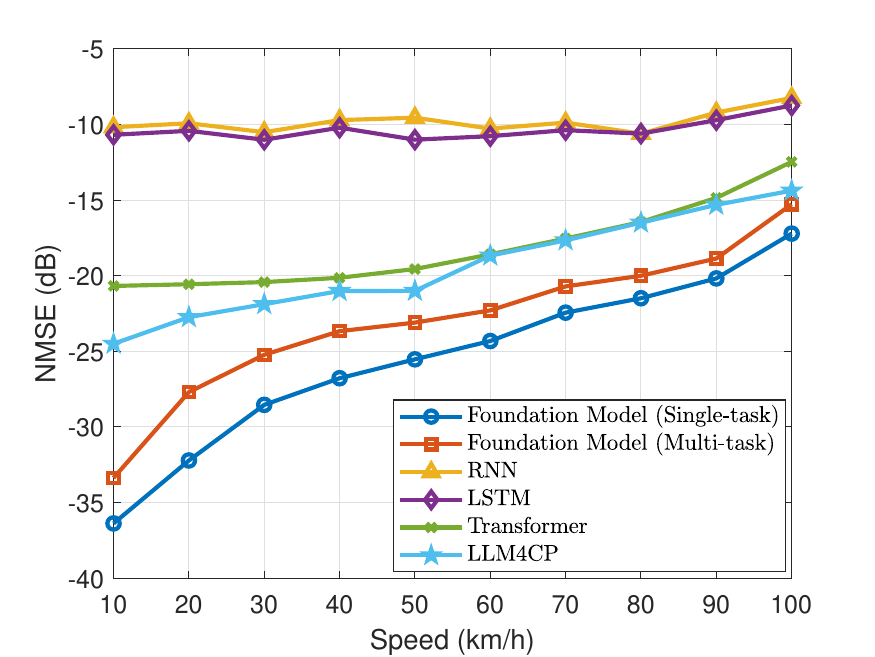}
	}
	\subfigure[$\Delta t = 50$  ms]{
		\includegraphics[width=0.48\textwidth]{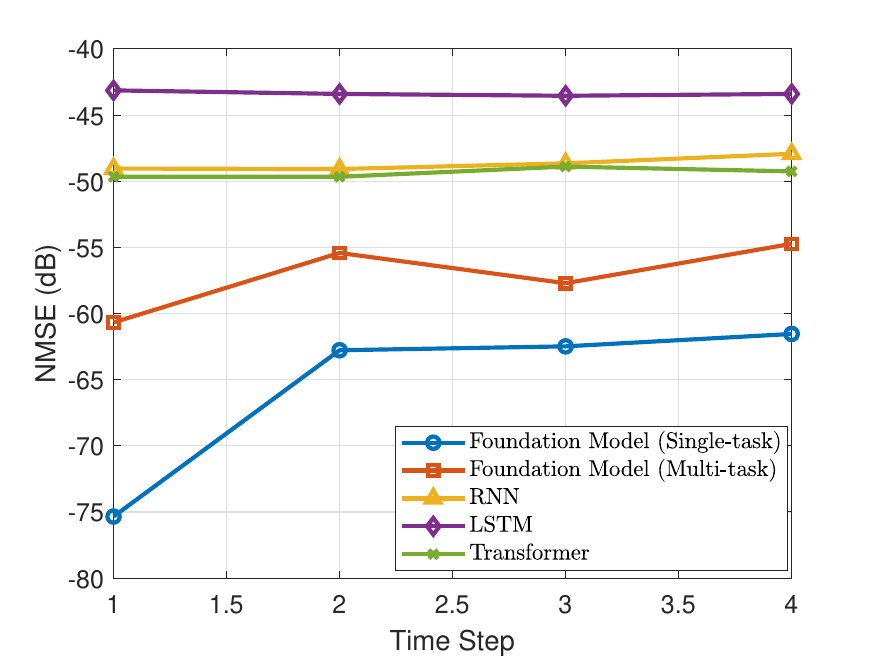} 
	}
	\caption{The NMSE performance of the proposed method compared with other baselines in the channel prediction task.}
    \label{fig:csi_base_result}
\end{figure*}

\subsection{Task I: MISO Channel Prediction}
\subsubsection{Baselines and Performance Metric}
To assess the effectiveness of the proposed multi-task foundation model on the channel prediction task, we compare it against several representative baseline models:

\begin{itemize}
    
    \item \textbf{RNN}: RNN \cite{jiang2019nnoverview} is a classical neural architecture commonly used for modeling sequential data, widely used in early channel prediction studies. In this work, we adopt a four-layer RNN to model temporal correlations in the channel data.

    \item \textbf{LSTM}:  LSTM \cite{jiang2020deep} improves upon standard RNNs by introducing memory cells and gating mechanisms, allowing better long-term dependency retention. We implement a four-layer LSTM for a fair comparison.

    \item \textbf{Transformer}: Following the approach in \cite{jiang2022transformer}, we implement a Transformer-based model that performs parallel prediction of future CSI. This architecture mitigates error propagation and improves inference efficiency.

    \item \textbf{LLM4CP}: LLM4CP \cite{liu2024llm4cp} represents an early attempt to adapt large pretrained language models, such as GPT-2, to the channel prediction domain. It exhibits enhanced robustness and accuracy under diverse wireless conditions.

    \item \textbf{Foundation Model (Single-task)}: For comparison, we also train our model only on the channel prediction dataset.
    
\end{itemize}
For quantitative evaluation, we adopt the NMSE as the primary metric, which is widely recognized for measuring the accuracy of channel prediction and is thus central to our performance analysis.

\subsubsection{Performance Analysis}

As shown in Fig.~\ref{fig:csi_base_result}(a), we compare the NMSE performance of our proposed foundation model with several baseline methods under varying user velocities, with a fixed time interval of $\Delta t = 0.5$ ms. The results indicate that NMSE performance consistently deteriorates across all models as the user's speed increases. This degradation stems from the rapid channel variations and shorter coherence times associated with higher mobility, which collectively increase the difficulty of accurate channel prediction. Notably, attention-based approaches---namely, Transformer, LLM4CP, and our foundation model---consistently outperform RNN and LSTM baselines, underscoring their effectiveness in capturing dynamic channel behavior. Our proposed foundation model consistently outperforms traditional baselines in both single-task and multi-task training settings, particularly at lower velocities. Additionally, it exceeds the performance of LLM4CP—a method based on adapting large language models—demonstrating the advantages of pretraining on large-scale time-series data tailored for communication tasks. The multi-task version of our foundation model exhibits slightly reduced accuracy compared to its single-task counterpart, which is expected due to its broader learning objective. Fig.~\ref{fig:csi_base_result}(b) further presents the NMSE performance across various prediction time steps with $\Delta t = 50$ ms. Since large-scale fading tends to be more temporally correlated and exhibits less fluctuation than small-scale fading, all evaluated models achieve markedly better prediction performance in this context. Importantly, our foundation model delivers substantial improvements over all baselines, reaffirming its powerful generalization and prediction capabilities in both short- and long-term channel modeling scenarios.

\begin{figure}[htp]
    \centering
    \includegraphics[width=0.5\textwidth]{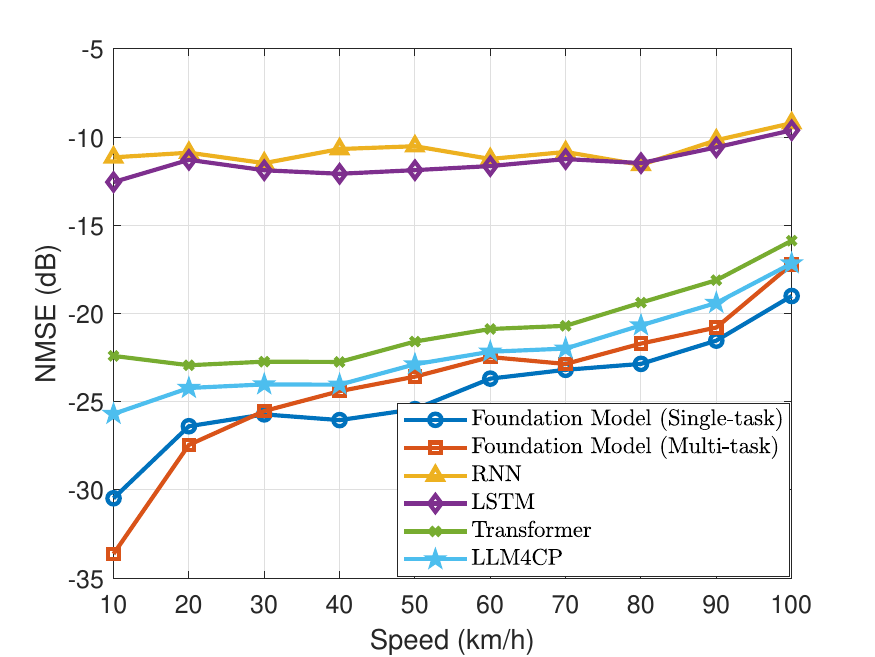}
    \caption{The zero-shot generalization performance in the D11 dataset of the channel prediction task.}
    \label{fig:csi_unseen_result}
\end{figure}

As shown in Fig.~\ref{fig:csi_unseen_result}, we evaluate the generalization of our proposed foundation models by directly applying the model trained in scenarios D1–D10 to an unseen scenario D11, without any additional training or adaptation. Our proposed foundation model consistently outperforms the baselines in terms of NMSE, underscoring its ability to generalize across different channel environments. Interestingly, the performance difference between the multi-task and single-task versions of our foundation model becomes less pronounced in this unseen setting. This observation suggests that multi-task training, by exposing the model to diverse conditions, can lead to improved generalization even when the test environment differs significantly from the training distribution.

\subsubsection{Downstream Task}
In addition, to evaluate the communication effectiveness of these methods, we compare the spectrum efficiency performance of foundation models with other baseline schemes in Table \ref{tab:se_comparison}. The communication SNR is set as 10 dB.
The maximum achievable spectrum efficiency is obtained with perfect CSI. The spectrum efficiency performance is calculated according to (\ref{eq:se}) and averaged over all test speeds. Our proposed foundation models, particularly the ``Foundation Model (Single-task)'', outperform traditional algorithms and approaches the maximum achievable spectrum efficiency.

\begin{table*}[ht]
\centering
\caption{Spectrum efficiency performance compared with baselines in the downstream task of channel prediction (Maximum achievable spectrum efficiency: 7.01 bps/Hz).}
\begin{tabular}{lcccccc}
\toprule
\textbf{Metric} & \textbf{RNN} & \textbf{LSTM} & \textbf{Transformer} & \textbf{LLM4CP} & \textbf{Foundation Model (Multi-task)} & \textbf{Foundation Model (Single-task)} \\
\midrule
$R_{\text{CP}}$ (bps/Hz) & 6.7634 & 6.7915 & 6.8508 & 6.8644 & 6.8701 & \textbf{6.8714} \\
\bottomrule
\end{tabular}
\label{tab:se_comparison}
\end{table*}

\subsection{Task II: Angle Prediction for ISAC Positioning}
\subsubsection{Baselines and Performance Metric}
To assess the effectiveness of the proposed multi-task foundation model, we compare it against several representative baseline models:
\begin{itemize}
    \item \textbf{CLRNet}: A convolutional long short-term memory (CLSTM) network specifically designed for ISAC scenarios \cite{liu2022isaclstm}. It combines CNNs for spatial feature extraction with LSTM units to capture temporal dynamics, and is tailored for predictive beamforming via AoA forecasting.
    \item \textbf{RNN}: A standard RNN baseline configured with four stacked layers to model temporal correlations in sequential angle data \cite{sherstinsky2020rnn}.
    \item  \textbf{Informer}: A Transformer-based architecture optimized for long sequence time-series forecasting \cite{zhou2021informer}. It incorporates sparse self-attention and a self-attention distillation mechanism to reduce computational complexity while maintaining accuracy. For angle prediction, we adopt a four-layer Informer configuration.
     \item \textbf{Foundation Model (Single-task)}: For comparison, we also train our model only on the angle prediction dataset.
\end{itemize}
For quantitative evaluation, we adopt the NMSE as the primary performance metric.
\subsubsection{Performance Analysis}

\begin{figure*}[htp]
    \centering
    \includegraphics[width=1\textwidth]{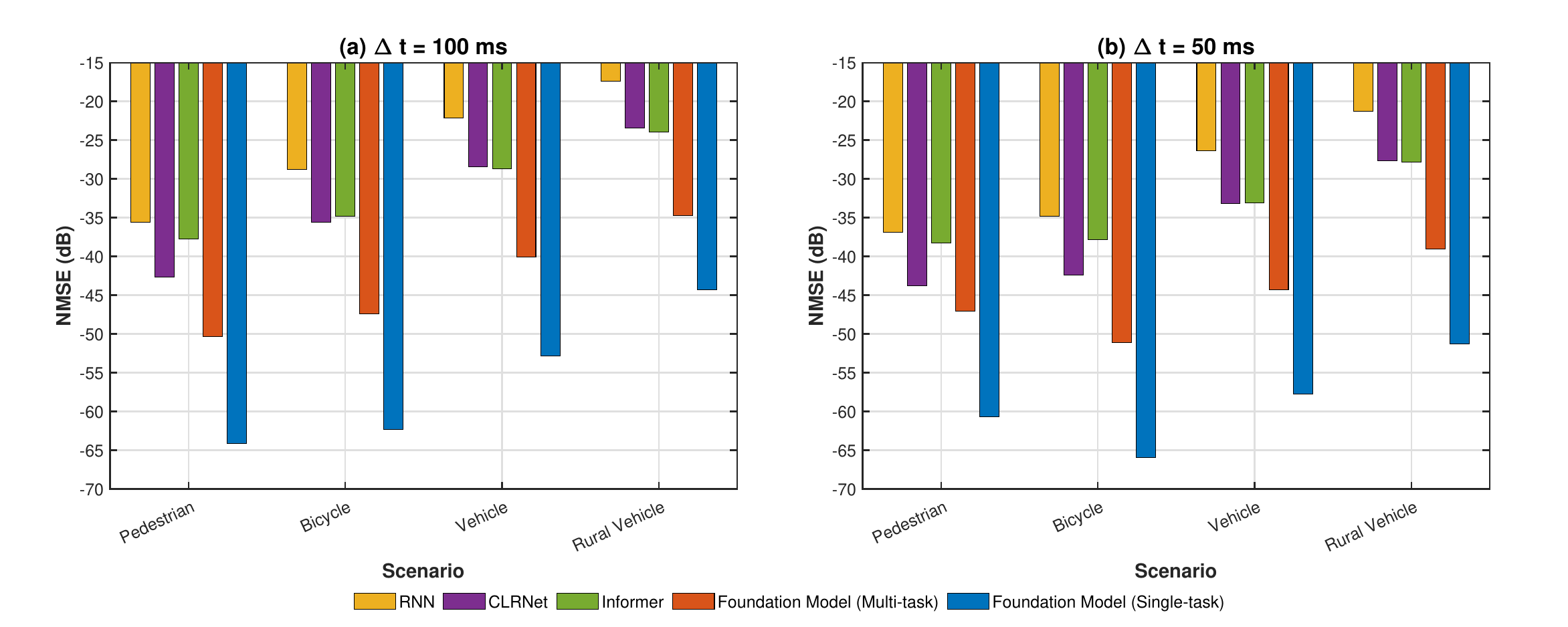}
    \caption{The NMSE performance of our proposed foundation models compared with other baselines in the angle prediction task across various scenarios.}
    \label{fig:trajectory_base}
\end{figure*}

We compare the NMSE performance of the foundation models with various baseline schemes across different scenarios, as shown in Fig.~\ref{fig:trajectory_base}. As the speeds of the four distinct motion scenarios increase from left to right, a consistent deterioration in performance is observed across all schemes. CLRNet, a model specifically designed for AoA prediction in ISAC, outperforms the naive RNN and Informer networks, both of which were not optimized for ISAC-related tasks. Upon comparing the performance of different schemes, it becomes evident that our proposed foundation models, whether in a single-task or multi-task configuration, significantly outperform the other baseline approaches. Additionally, the single-task model, trained on a specific task, demonstrates a remarkable performance improvement due to its tailored optimization. A more detailed examination of Fig.~\ref{fig:trajectory_base}(a) and Fig.~\ref{fig:trajectory_base}(b) shows that, in low-speed scenarios (such as Pedestrian and Bicycle), the effect of the prediction interval $\Delta t$ on the performance of all schemes is minimal. In contrast, in high-speed scenarios (such as Vehicle and Rural Vehicle), performance declines as $\Delta t$ increases. This degradation is attributed to the fact that, in high-speed environments, longer prediction intervals result in larger object displacements, which makes it more difficult to predict the AoA accurately. As a result, larger values of $\Delta t$ lead to a more significant reduction in performance.

\begin{figure*}[htp]
    \centering
    \includegraphics[width=1\textwidth]{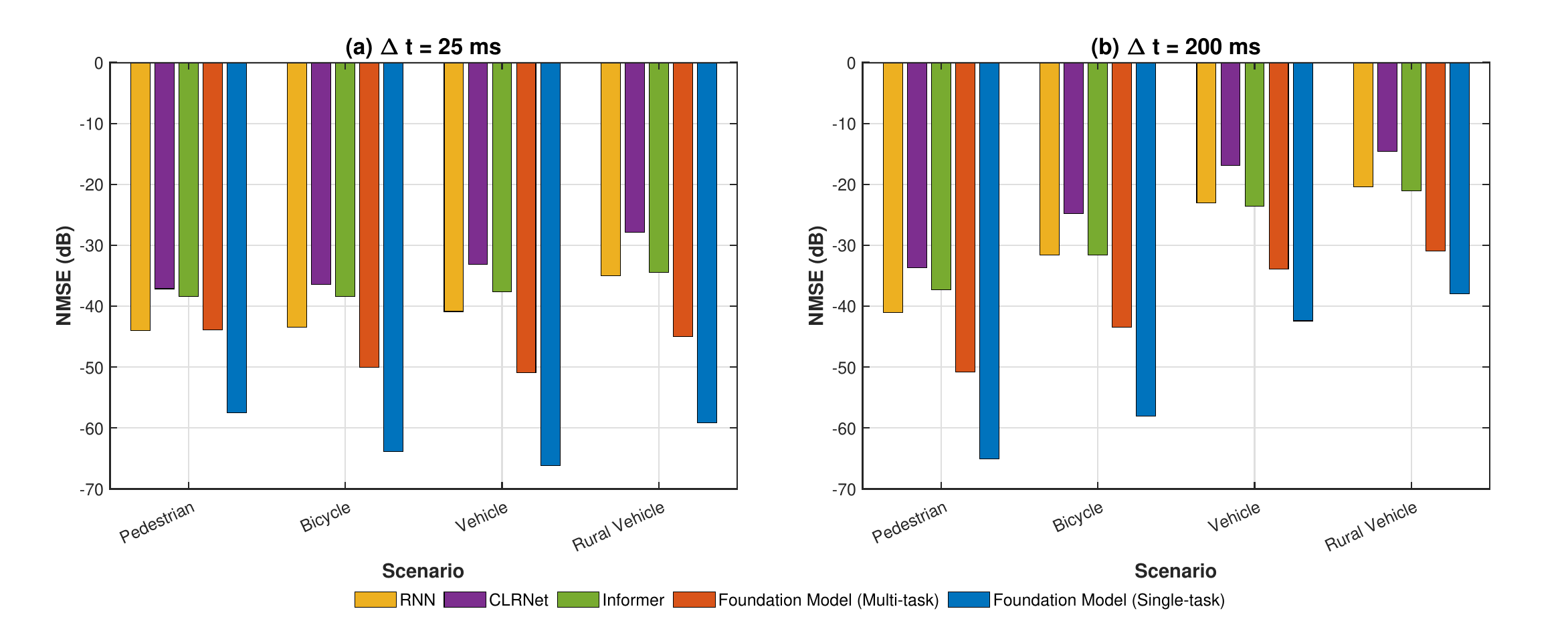}
    \caption{The zero-shot generalization performance of our proposed foundation models compared with other baselines in the angle prediction task across various scenarios with unseen prediction intervals.}
    \label{fig:trajectory_unseen}
\end{figure*}

We also compare the NMSE performance of the foundation models with various baseline schemes across different scenarios with unseen prediction intervals, as shown in Fig.~\ref{fig:trajectory_unseen}. We train all models at $\Delta t = 50~\text{or}~100$ ms and directly test them at $\Delta t = 25~\text{or}~ 200$ ms without retraining. From Fig.~\ref{fig:trajectory_unseen}(a), it is evident that our proposed foundation model, particularly the single-task model, outperforms other baselines across various scenarios, demonstrating superior generalization capabilities. As shown in Fig.~\ref{fig:trajectory_unseen}(b), in high-speed scenarios, the further increase in $\Delta t$ exacerbates the difficulty of prediction, leading to a performance degradation across all schemes. However, the generalization strength of our proposed foundation model becomes even more pronounced, as it maintains relatively better performance compared to other approaches, highlighting its robustness when faced with unseen prediction intervals.

\subsubsection{Downstream Task}
To evaluate the effectiveness of the predictive beamforming methods, we compare the downlink spectrum efficiency, as calculated in (\ref{eq:sum_rate_isac}), of our proposed foundation models with various baseline schemes, as shown in Table \ref{tab:sum_rate_comparison}. In the simulations, we set $K = 8$ single-antenna users and $N_t = 4,096$. A path loss model is used to characterize large-scale fading, expressed as $\alpha_{k,n} = \alpha_0 (d_{k,n}/d_0)^{-\zeta}$, where $\alpha_0 = -65$ dB, $d_0 = 1$ m, and $\zeta = 3$ \cite{niu2015survey}. Additionally, the noise variance is set to $\sigma_k^2 = -80$ dBm for all users, and the transmit power for the $k$-th user is $p_{k,n} = P_{\text{total}}/K$, with $P_{\text{total}} = 20$ dBm representing the total transmit power. As shown in Table \ref{tab:sum_rate_comparison}, our proposed foundation models significantly outperform other baseline approaches in the predictive beamforming task. This superior performance can be attributed to the high precision required for beamforming in large-scale MIMO systems, where the beams are very narrow. The accuracy of angle predictions is crucial in this context, and our foundation models, with their enhanced capabilities in angle prediction, far surpass the baselines. Consequently, this leads to more precise predictive beamforming, resulting in a considerable increase in the downlink spectrum efficiency. Thus, the exceptional performance of our foundation models in this downstream task highlights their significant potential in improving system efficiency in practical communication scenarios.

\begin{table*}[ht]
\centering
\caption{Average achievable spectrum efficiency of foundation models compared with baselines in the downstream task of angle prediction  (Maximum achievable spectrum efficiency: 54.67 bps/Hz).}
\begin{tabular}{lcccccc}
\toprule
\textbf{Metric} & \textbf{RNN} & \textbf{CLRNet} & \textbf{Informer} & \textbf{Foundation Model (Multi-task)} & \textbf{Foundation Model (Single-task)} \\
\midrule
$R_{\text{AP}}$ (bps/Hz)   & 13.67 & 16.87 & 15.51  & 33.88 & \textbf{39.93} \\
\bottomrule
\end{tabular}
\label{tab:sum_rate_comparison}
\end{table*}

\subsection{Task III: Traffic Prediction} 
\subsubsection{Baselines and Performance Metric}
To assess the effectiveness of the proposed multi-task foundation model, we compare it against several representative baseline models:
\begin{itemize}
    \item \textbf{LSTM}: LSTM \cite{chen2017lstmtraffic} is a common method for deep learning to predict time-series data. We utilize a four-layer LSTM configuration for this task.
    \item  \textbf{STGCN}: The space–time graph convolutional network (STGCN) \cite{yu2017STGCN} is the first article that integrates temporal and spatial features through GNNs, and is widely used for comparison.
    \item  \textbf{ASTGNN}: In \cite{guo2019ASTGNN}, the authors propose an attention-based spatial-temporal graph neural network (ASTGNN) for traffic forecasting, which relies on the traffic data from neighboring BSs.
     \item \textbf{Foundation Model (Single-task)}: For comparison, we also train our model only on the traffic prediction dataset.
\end{itemize}
For quantitative evaluation, we adopt the NMSE as the primary performance metric.
\subsubsection{Performance Analysis}

\begin{figure}[htp]
    \centering
    \includegraphics[width=0.5\textwidth]{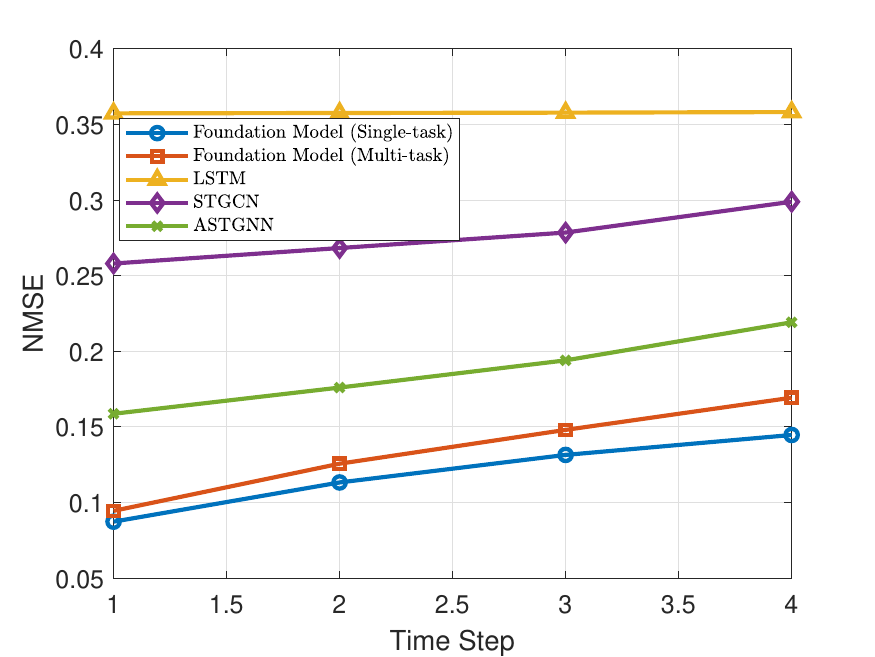}
    \caption{The NMSE performance of the foundation models compared with baselines in the traffic prediction task across time steps.}
    \label{fig:traffic}
\end{figure}

As shown in Fig.~\ref{fig:traffic}, we compare the NMSE performance of the foundation models with the baselines in all time steps. In contrast to the naive LSTM prediction approach, both STGCN and ASTGNN utilize GNNs to model spatial dependencies, which results in improved performance. Furthermore, ASTGNN enhances the performance of STGCN by incorporating attention mechanisms to capture temporal dependencies more effectively. However, our proposed foundation models, which do not rely on the traffic data from neighboring BSs, outperform ASTGNN. This is due to the long context length employed by our models, which enables accurate traffic prediction based on historical data alone. This ability is especially significant in practical applications, as obtaining traffic information from other BSs may be impractical or impossible due to privacy concerns.

\subsection{Zero-Shot and Scaling Analysis}
In the previous subsections, we evaluated the zero-shot generalization ability of our foundation model across different configurations within the same task. In this subsection, we take a further step and investigate the model’s zero-shot performance on an unseen task, thereby assessing its capacity for out-of-task generalization. Specifically, we consider a new task: predicting the time-delay in an ISAC system, which corresponds to estimating the physical distance between a user and the BS, i.e. $\nu_{k,t}$ in (\ref{eq:isac}). Structurally, this task resembles previous angle prediction tasks, such as the one described in (\ref{eq:angle_prediction}).  In the following results, we compare the NMSE across time steps for various models to evaluate their performance on the unseen delay prediction task. The RNN, CLRNet, Informer, and Foundation Model (Single-task) are all trained and evaluated under a full-shot setting, where the target task (time-delay prediction) is directly included during training. These baseline schemes follow the same training configuration as Task II, ensuring a fair and consistent comparison.  In contrast, the Foundation Model (1-task), (2-task), and (3-task) variants are trained only on Task I, Task I \& Task II, and Task I, Task II \& Task III, respectively, without access to the delay prediction task during training. These models are then directly evaluated on the unseen task in a strict zero-shot setting. 
\begin{figure}[htp]
    \centering
    \includegraphics[width=0.5\textwidth]{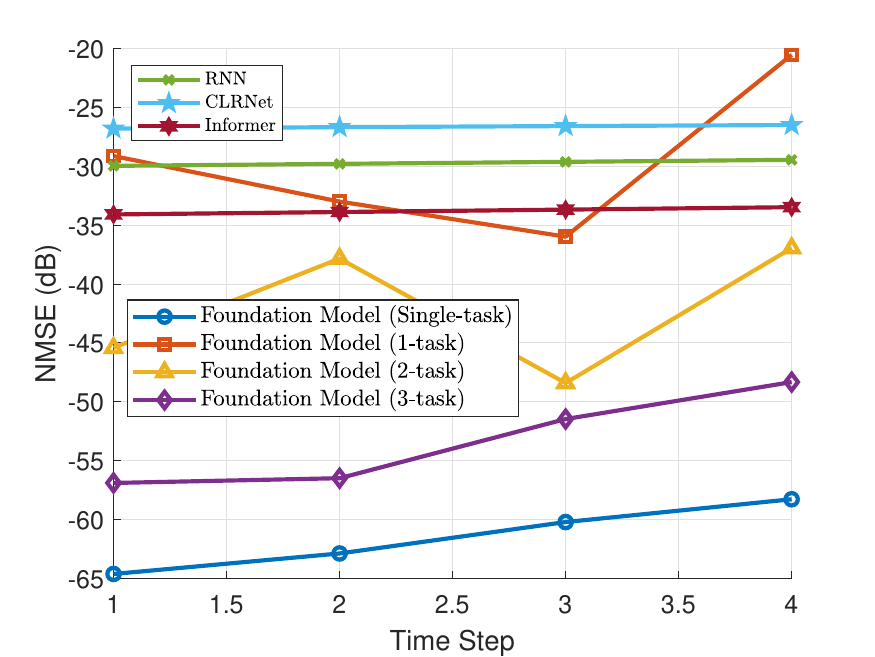}
    \caption{The zero-shot generalization performance in the time-delay prediction task.}
    \label{fig:unseen_task}
\end{figure}

\begin{table*}[htbp]
\centering
\caption{Ablation experiments on different modules of the foundation models}
\begin{tabular}{lcccc}
\toprule
\textbf{Task} & \textbf{Channel Prediction} & \textbf{Channel Prediction} & \textbf{AoA Prediction} & \textbf{Traffic Prediction} \\
                & ($\Delta t = 0.5$ ms) & ($\Delta t = 50$ ms) & ($\Delta t = 50/100$ ms) & ($\Delta t = 1$ h) \\
\midrule
Base & 0.00823 & $\boldsymbol{2.20 \times 10^{-6}}$ & $\boldsymbol{8.15 \times 10^{-5}}$ & \textbf{0.134} \\
w/o Positional Encoding & 0.00873 & $2.39 \times 10^{-6}$ & $9.10 \times 10^{-5}$ & 0.138 \\
w/o Granularity Encoding & 0.0236 & $ 2.97 \times 10^{-6} $ & $ 1.96 \times 10^{-4} $ & 0.211 \\
w/o Patch & \textbf{0.00803} & $ 2.28 \times 10^{-6} $ & $ 8.50 \times 10^{-5} $ & 0.153 \\
\bottomrule
\end{tabular}
\label{tab:ablation}
\end{table*}

As shown in Fig.~\ref{fig:unseen_task}, the proposed foundation model demonstrates strong generalization ability when trained on multiple tasks. Notably, its zero-shot performance on the unseen time-delay prediction task significantly outperforms traditional baselines such as RNN, CLRNet, and Informer. Furthermore, the model trained on all three tasks (i.e., foundation model (3-task)) achieves performance that closely approaches that of the fully supervised foundation model (Single-task), which is directly trained on the target task. In addition, comparing the Foundation Model (1-task), (2-task), and (3-task) variants reveals a clear trend: pretraining on a larger number of diverse tasks consistently improves the model’s zero-shot generalization capability. This confirms the benefit of task diversity in training, and highlights the scalability of the foundation model framework in addressing unseen tasks without requiring retraining or finetuning.

\begin{figure}[htp]
    \centering
    \includegraphics[width=0.5\textwidth]{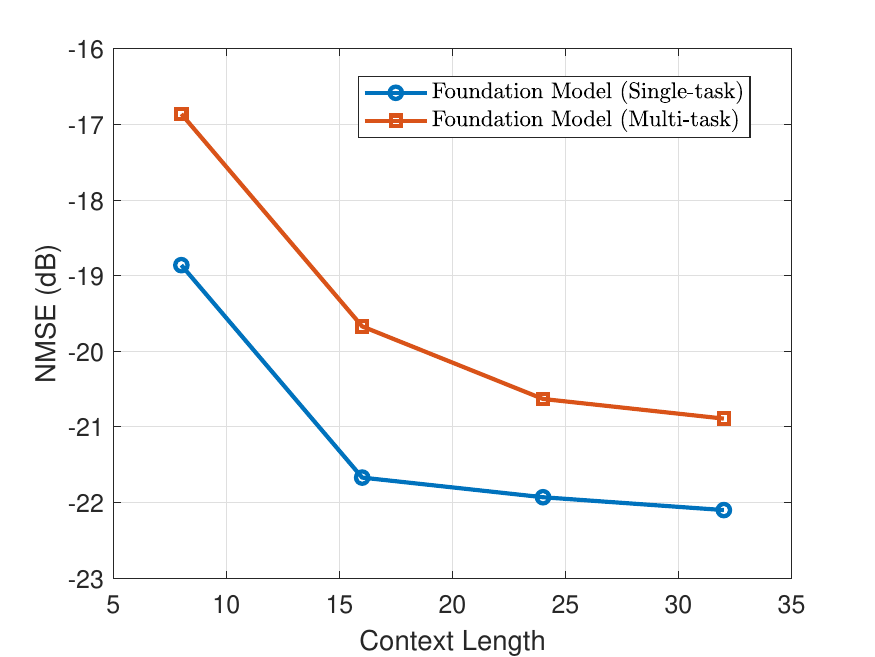}
    \caption{The NMSE performance of the foundation models for the channel prediction task across different context lengths, $L$.}
    \label{fig:context_length}
\end{figure}
We further evaluate the NMSE performance of the foundation models for the channel prediction task across different context lengths, $L$. As illustrated in Fig.~\ref{fig:context_length}, our proposed foundation model, which incorporates a masking strategy during pretraining, is capable of accommodating any input within the range of 1 to 32 for the channel prediction task ($\Delta t = 0.5 \, \text{ms}$). The NMSE performance of the foundation models improves with the increasing context length. However, once the context length approaches 32, the performance improvement becomes marginal. This behavior can be attributed to the inherent nature of the channel prediction task, where only the channels within the coherence time exhibit significant correlations. Consequently, an unrestricted increase in context length does not lead to a continuous performance improvement, as the additional information beyond the coherence time becomes progressively less relevant for accurate channel prediction.

\begin{figure}[htp]
    \centering
    \includegraphics[width=0.5\textwidth]{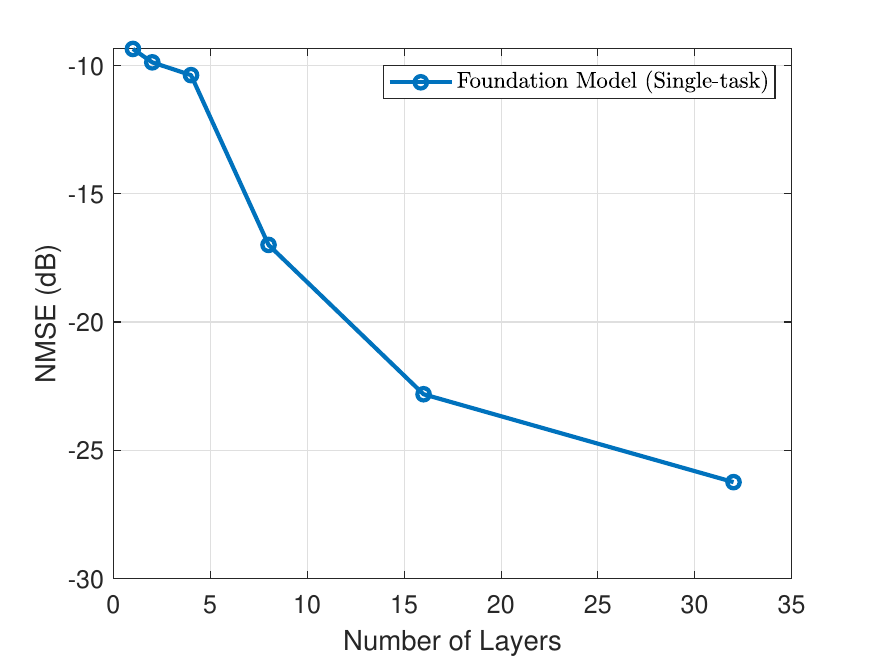}
    \caption{The NMSE performance of the foundation model for the channel prediction task under different numbers of Transformer layers.}
    \label{fig:scaling_law}
\end{figure} 

As illustrated in Fig.~\ref{fig:scaling_law}, we evaluate the NMSE performance of the foundation model for the channel prediction task under different numbers of Transformer layers. We adjust the number of Transformer layers to study the impact of the foundation models' parameters on performance. When the number of Transformer layers is fewer than 4, the performance is suboptimal. However, once the number of Transformer layers exceeds 8, the performance experiences a dramatic increase. This suggests that the foundation model benefits significantly from deeper Transformer architectures, where the increased depth allows for better feature extraction and enhanced representation learning, leading to substantial improvements in the channel prediction task.

\subsection{Ablation Studies} \label{sec:ablation studies}

As shown in the Table~\ref{tab:ablation}, we conducted ablation experiments on different modules of the foundation models, with the performance metric being the NMSE on all test datasets for each task. It is evident that removing the positional encoding (w/o Positional Encoding) leads to a performance degradation. This is because positional encoding plays a crucial role in guiding the model for sequence prediction, providing temporal context to the input data. Furthermore, the removal of granularity encoding (w/o Granularity Encoding) results in a more significant performance decline, particularly for the tasks of channel prediction ($\Delta t = 0.5$ ms) and traffic prediction ($\Delta t = 1$ h). This can be attributed that granularity encoding is essential for distinguishing between tasks with varying sampling interval characteristics. Since the channel prediction ($\Delta t = 0.5$ ms) and traffic prediction ($\Delta t = 1$ h) tasks exhibit the largest sampling interval differences, the absence of granularity encoding has a particularly pronounced effect on these tasks. Finally, the removal of the patch module (w/o Patch) has the greatest impact on Traffic Prediction. This is due to the long context length required for traffic prediction, which necessitates the use of patches to aggregate information over extended time periods. Without the patch module, the foundation models struggle to effectively process long-range dependencies, resulting in a notable performance drop for this task.

\section{Conclusions}\label{sec:conclusions}

In this paper, we have introduced a unified multi-task foundation model for time-series prediction in wireless networks, designed to accommodate diverse prediction intervals and heterogeneous tasks. By integrating univariate decomposition, granularity encoding, and a causal Transformer backbone, our framework unifies multiple prediction tasks, such as channel prediction, angle prediction, and traffic prediction, into a single scalable model. The use of patch masking during pretraining further enables the model to support arbitrary input lengths. Extensive experimental results demonstrate that the foundation model not only exhibits strong performance across all evaluated tasks, but also exhibits excellent generalization to unseen scenarios. Notably, its zero-shot performance on previously untrained tasks even surpasses the fully supervised performance of conventional baselines. Furthermore, downstream evaluations based on the model's predictions confirm its practical utility, consistently outperforming traditional task-specific methods. These findings suggest that the proposed foundation model effectively captures the underlying principles of time-series forecasting and offers a promising direction for the development of intelligent, general-purpose predictive models in wireless communications.


{\small
\bibliographystyle{IEEEtran}  
\bibliography{journal_dc.bib}  
}

\end{document}